\def\year{2021}\relax
\documentclass[letterpaper]{article} 
\usepackage{aaai21}  
\usepackage{times}  
\usepackage{helvet} 
\usepackage{courier}  
\usepackage[hyphens]{url}  
\usepackage{graphicx} 
\usepackage{natbib}  
\usepackage{caption} 
\usepackage{subcaption}
\usepackage{bm}
\usepackage{amsmath}

\usepackage{booktabs}

\title{Probabilistic Modeling of Human Teams to Infer False Beliefs}
\author {
    Paulo Soares, 
    Adarsh Pyarelal, 
    Kobus Barnard 
    \\
}
\affiliations {
    University of Arizona\\
    paulosoares@email.arizona.edu, adarsh@arizona.edu, kobus@arizona.edu
}
\begin{document}

\maketitle

\begin{abstract}

    We develop a probabilistic graphical model (PGM) for artificially
    intelligent (AI) agents to infer human beliefs during a simulated urban
    search and rescue (USAR) scenario executed in a Minecraft environment with
    a team of three players. The PGM approach makes observable states and
    actions explicit, as well as beliefs and intentions grounded by evidence
    about what players see and do over time. This approach also supports
    inferring the effect of interventions, which are vital if AI agents are to
    assist human teams. The experiment incorporates manipulations of players'
    knowledge, and the virtual Minecraft-based testbed provides access to
    several streams of information, including the objects in the players' field
    of view. The participants are equipped with a set of marker blocks that can
    be placed near room entrances to signal the presence or absence of victims
    in the rooms to their teammates. In each team, one of the members is given
    a different legend for the markers than the other two, which may mislead
    them about the state of the rooms; that is, they will hold a false belief.
    We extend previous works in this field by introducing ToMCAT, an AI agent
    that can reason about individual and shared mental states. We find that the
    players' behaviors are affected by what they see in their in-game field of
    view, their beliefs about the meaning of the markers, and their beliefs
    about which meaning the team decided to adopt. In addition, we show that
    ToMCAT's beliefs are consistent with the players' actions and that it can
    infer false beliefs with accuracy significantly better than chance and
    comparable to inferences made by human observers.

\end{abstract}

\section{Introduction}

Human-machine teaming has emerged in the last two decades as a way to describe
the ideal way for humans to interact with artificially intelligent (AI) agents
\cite{Urlings.ea:2001, Madni.ea:2018}, though not without some pushback
\cite{Groom.ea:2007}. At least one study has shown that humans are somewhat
open to the idea of AI agents as teammates, rather than tools
\cite{Lyons.ea:2018}. In this paper, we conceptualize human-machine teams not
as exact analogues of human-human teams, but rather collaborations with the
following features: autonomy, shared goals and knowledge, interdependence, and
complementary responsibilities.

There is a growing consensus that for effective human-machine teaming, AI
agents should be capable of building and maintaining models of their human
teammates \cite{Laird.ea:2019}. In other words, AI agents should develop a
`theory of mind' for their human teammates. We deviate slightly from the original
definition of theory of mind \cite{Premack.ea:1978} and the way it is currently
used in the literature. Rather than using the term to mean the ability to
recognize that other humans have distinct mental states, in our work, we
attempt to build an agent that \emph{assumes} the existence of distinct human
mental states and tries to infer them.

Unlike previous works that have focused on computational theory
of mind for single humans in complex environments \cite{Oguntola.ea:2021}, or
teams of artificial agents performing relatively simple tasks
\cite{Shum.ea:2019}, our work studies teams of humans collaboratively
performing tasks in a complex environment.

Research on human team performance shows that teams with strong `shared mental
models' perform better \cite{Cannon-Bowers.ea:1993}. We posit that one of the
potentially helpful actions that can be performed by an agent with theory of
mind is to detect when the mental models of its human teammates are misaligned
and intervene to align them, thus strengthening the team's shared mental model
and improve the team's performance. 

To this end, we develop and evaluate a probabilistic graphical model (PGM) for
inferring human belief states in a complex multi-participant simulated urban
search and rescue task.

\noindent Notably, our approach has a number of desirable features:

\noindent \textbf{(1)} Our model is \emph{interpretable}, with latent variable
nodes corresponding to high-level constructs that can easily interface with
social science theories.

\noindent \textbf{(2)} Our model explicitly represents the \emph{team's shared
mental model} as a latent node, separate from the individual mental models.

\noindent \textbf{(3)} Our algorithm is \emph{online}. Thus, the agent's models
of the individual human teammates and the team's shared mental model are
updated in real time, making it possible for the agent to intervene to correct
misaligned mental models as soon as they are detected.

\section{Related Work}

Recent works have addressed designing AI agents that can establish a theory of
human minds using a variety of approaches. \citet{Rabinowitz:2018} implemented
an AI observer that used meta-learning over several policy-driven simulated
agents to learn strong priors for the agents' behaviors and accurately inferred
subsequent actions on an unseen population. Their observer was also able to
identify potential changes of desires in a hand-crafted false-belief scenario.
However, experiments in this work were all conducted in small gridworld
settings, with simulated agents, rather than human subjects. Moreover, the
beliefs and desires were never directly measured. Rather, they were
implicitly assumed given the predicted agents' actions.  

An AI observer using neural sequences in a more complex, multi-goal scenario
also showed promising results on predicting strategies \cite{Jain.ea:2020}.
Simulated agents followed a non-optimal trajectory based on behaviors
collected from human subjects acting in the same environment. The AI observer
was then trained on the behavior of these faux human agents and surpassed human
observer predictions. However, this work did not incorporate beliefs and
desires and did not address false-belief assumptions.

Whereas discriminative models can achieve significant performance levels in
some scenarios, it is difficult to explain why the agents made a specific
prediction. \citet{Oguntola.ea:2021} attempt to address this issue by adding
explicit components such as human intents and concepts to these models to
increase their interpretability.

Interpretability is crucial for agents that aim to develop a theory of the
human mind, especially if these agents are to assist humans. It can explain the
reasons why an AI agent arrived at a particular belief, or performed a
particular intervention. We can compare the agents' cognitive process to the
mechanisms involved in the humans' perception of the world and use this as a
qualitative measure of the plausibility of the agent's theories. Bayesian
models are a natural framework to work with explicit representations of mental
states and have already showed an ability to infer beliefs and desires, closely
aligning with predictions made by human observers \cite{Baker:2017}. 

This approach has also shown to be a good choice to model non-optimal behavior
in the context of planning \cite{Zhi-Xuan.ea:2020, Alanqary.ea:2021}. Humans do
not always act optimally and might behave differently from an ideally rational
agent. A core concept in a Bayesian framework is the manipulation of
uncertainty, which, in turn, can be a solution to deal with such a problem.

A general limitation of most computational theory-of-mind related works is that
they do not address the problem in a scenario in which more than one individual
is involved. Therefore, they cannot explain changes in human behavior when they
have to coordinate with each other. Developing a theory of a `team mind' is a
more challenging task, and might require revisiting some premises used for a
single human scenario.

\section{Mission Design}

\begin{figure}
	\centering
	\begin{subfigure}{\linewidth}
	    \centering
		\includegraphics[width=0.8\linewidth]{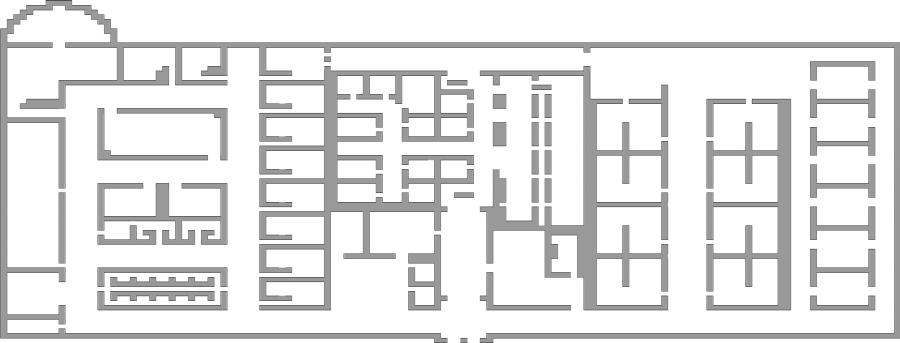}
		\caption{2D view of the building floor plan.}
		\label{fig:2d_map}		
	\end{subfigure}
	\begin{subfigure}{.5\textwidth}
	    \vspace{1em}
		\centering
		\includegraphics[width=0.8\linewidth]{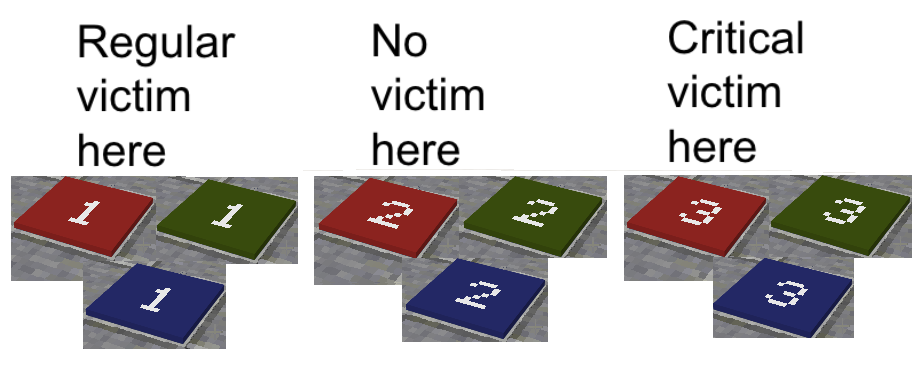}
		\caption{Marker meaning legend B as displayed on the screen.}
		\label{fig:marker_legend}
	\end{subfigure}
	\caption{Floor map of the simulated environment and marker meaning legend B
    that one of the three players sees on the screen.}
	\label{fig:mission_design}
\end{figure}

The experiments consist of simulated urban search and rescue missions executed
in a virtual world developed within Minecraft. The environment~(see
Figure~\ref{fig:2d_map}) is a collapsed office building. Teams of three humans
must search the building for victims to triage. These victims may be located
inside rooms or hidden under piles of rubble. The team earns a certain number
of points for each victim that is triaged (`critical', i.e., severely injured
victims, are worth more points than `regular' victims, and require all three
teammates to be present to triage them). The objective of the mission is
to maximize the number of points earned by the team within a 15 minute time
limit.

Players can choose among the following three possible roles, each of which has
an associated tool with limited durability: (i) search specialists can use a
stretcher to move regular victims from one location to another, (ii) engineers
can use a hammer to clear rubble piles to reveal victims or open blocked areas,
and (iii) medics can use a medkit to triage victims. There are no restrictions
on the role each player can assume, and more than one player can take on the
same role simultaneously if the team decides that would be a good strategy.

Roles are chosen at the beginning of the mission. A player can change their
role by returning to a `base' area, and also replenish their tools
(stretcher/hammer/medkit), which have a limited number of uses. In this way, a
time cost is associated with changing roles and using tools. Also, the players have access to the tool age at all times. Therefore, they might decide to replenish their tools before they become unusable.

Players communicate via voice chat with each other, through which they can
coordinate their strategies and actions. The players can choose to place marker
blocks labeled as either 1, 2, or 3 that they can use all around the building
to track information relevant to the team (e.g., rooms with victims that need
to be triaged).  There is no limit on how many marker blocks players can use.
Before the mission starts, each player receives a legend describing the meaning
of the marker blocks. Two players receive legend A, while the third receives a
different version, legend B, in which the meanings of markers 1 and 2 are
swapped. This generates a false belief on the part of the player who gets
legend B about the meaning of the marker blocks. The content of legends A and B
are given in Table \ref{tab:marker_legends}.  The participants are not told
that their legend might be different from that of their teammates. The legend
assignment conditions has three equally likely cases corresponding to which
player received legend B. 

\begin{table}
    \small
    \begin{tabular}{lll}
        \toprule
        Marker & Legend A & Legend B\\\midrule
        1 & No victim here & Regular victim(s) here\\
        2 & Regular victim(s) here & No victim here\\
        3 & Critical victim(s) here & Critical victim(s) here\\
        \bottomrule
    \end{tabular}
    \caption{The different marker legends received by the teammates in the experiment.}
    \label{tab:marker_legends}
\end{table}

\section{Probabilistic Model of Minds}

We propose an AI agent called ToMCAT (Theory of Mind-based Cognitive
Architecture for Teams) to infer each player's beliefs about marker meaning and
the collective team agreement on marker meaning. Specifically, we infer which
player received legend B, which we can validate.

ToMCAT reasons about which meaning the players are using for the markers as they place them in the environment or take action after markers have been in their in-game field of view. The agent represents its theory of the players' and team's minds using a probabilistic model for its beliefs about players' individual and collective mental states. Latent random variables represent these states, and the agent updates its beliefs by doing inference over them, grounded by evidence about what the players see and do over time.

We choose a probabilistic approach for our agent to explicitly represent the players' beliefs and states. This representation allows us to understand how the agent reasons about the players' beliefs and facilitates evaluating the effect of potential interventions for assisting human teams. Further, we can inspect the representation and evaluate whether the model is reasonably achieving predictions. 

In the instrumented mission domain, we hypothesize that players will start by placing markers consistent with the original legend they received. Moreover, the team might collectively follow a specific legend; this will explain situations in which players place markers that differ in meaning from the players' original legends. 

We use as evidence (i) observations of victims appearing in the players'
in-game field of view and (ii) placement of marker blocks 1 and 2. We captured the players' in-game view using a resolution of 16:9, which spans the field of view in 70$^{\circ}$ vertically (default setting of Minecraft's camera) and 125$^{\circ}$
horizontally (from the resolution ratio). We do not use placements of marker block 3 as evidence because
the meaning of this marker is the same for both legends, and thus we can
reasonably assume that everyone has this belief. Additionally, we simplify the
world by making the following assumptions: 

\begin{itemize}

    \item The team adopted legend does not change over time. Therefore, we can
        represent it as a single-time variable in the model. This is different
        from the probability changing over time as the agent observes more
        evidence. The agent's belief about the true single value of this latent
        state will change over time.  

    \item Players will use the markers according to the definitions provided in the legends. The agent might fail to infer the correct original assignment if the player used a meaning for the marker that is not consistent with legends A and B.  (e.g., if a player chose to use markers to keep track of rooms with rubble that needed to be cleared).

    \item The players will perceive any victim in their in-game field of view,
        that is, we assume that every time a victim is visible on a player's
        computer screen, the player will notice them. Victims are hidden inside rooms or under piles of rubble. Therefore, the players will invariably be close to the victims when they find them, which makes this assumption plausible.

    \item This perception will last five seconds on average, which means that
        if players place markers a long time after noticing victims, the agent
        will interpret that the players are using the markers to inform others
        that there is no victim nearby. 
        
    \item Given the same victim perception, original and team legend, players will behave similarly. This implies that the parameters of the distributions in the model can be shared among the players. 
\end{itemize}

Formally, the agent holds: (i) a belief about the original marker meaning
legend that each participant received, $O^{(i)}$ - where $i \in \{1,2,3\}$
represents a specific player - (ii) a belief about which meaning the team
decided to adopt collectively, $T$, (iii) a belief about whether the players
perceived a victim in their in-game field of view, $P^{(i)}_t $, at a given
time step $t$ (each time step is one second in length), and (iv) a belief about the
original legend assignment, $A$. This last latent state combines the beliefs
about the individual players, $O^{(i)}$, embedding the restriction that only
one of them received legend B. The model is grounded by victims in the players'
in-game field of view, $F^{(i)}_t$, and markers, $M^{(i)}_t$, they place at any
time step (see \figurename \ \ref{fig:pgm}). 

Hence, let $\tau$ be the last time step in a trial. The model defines the following joint probability over latent beliefs and observations:

\begin{figure}[!tb]
  \centering
  \includegraphics[width=\linewidth]{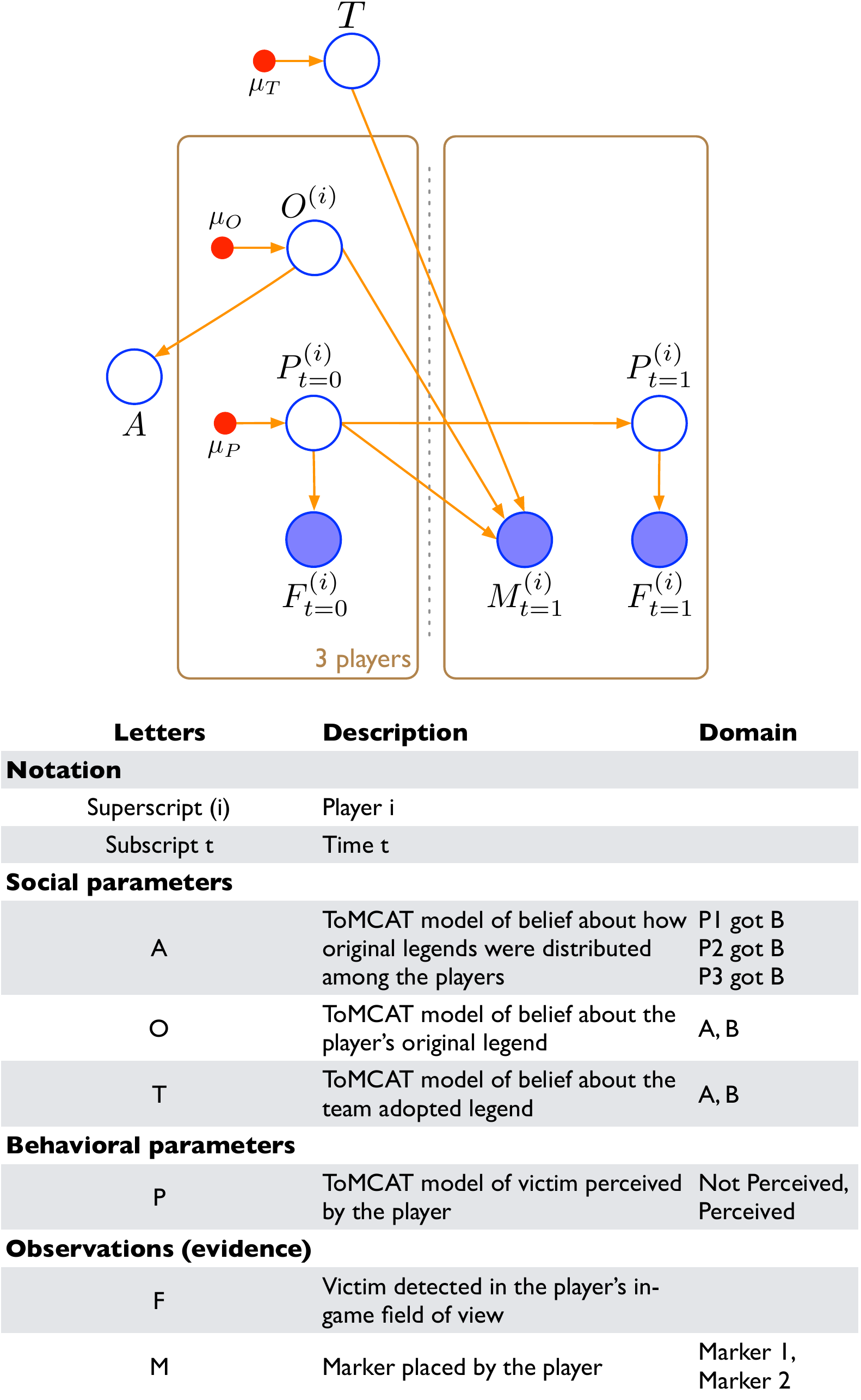}
  \caption{%
      Players' mental states represented as a dynamic Bayes net (DBN). The
      model encodes the joint distribution of the agent's beliefs, grounded by
      what the players see and do.           
  }
  \label{fig:pgm}
\end{figure}

\begin{align}
\begin{split}
p&(A, T, O^{(i)}, P^{(i)}_{0\dots \tau}, F^{(i)}_{0\dots \tau}, M^{(i)}_{1\dots \tau}) = p(A|O^{(1)},O^{(2)},O^{(3)})\\
	&p(T)\prod_{i \in \{1,2,3\}}p(O^{(i)})p(P^{(i)}_0)p(F^{(i)}_0|P^{(i)}_0)\\&\prod_{t=1}^{\tau}p(M^{(i)}_t|P^{(i)}_{t-1},O^{(i)}, T)
    p(F^{(i)}_t|P^{(i)}_t)p(P^{(i)}_t|P^{(i)}_{t-1}) \vphantom{\prod_{i \in \{1,2,3\}}}
\end{split}
\label{eq:joint_probability}
\end{align}

Next, let $I_\text{Perceive}$ be an indicator function of victim perception. We
define the priors and conditional distributions as follows:

\begin{align}
\begin{split}
		\mu_T, \mu_O, \mu_P &= 0.5\\
		T &\sim \text{Binomial}(\mu_T)\\ 		
		O^{(i)} &\sim \text{Binomial}(\mu_O)\\ 		
		P^{(i)}_0 &\sim \text{Binomial}(\mu_P)\\ 
		P^{(i)}_t|P^{(i)}_{t-1} = \text{Not Perceived} &\sim \text{Categorical}([0.8, 0.2]) \\	
		P^{(i)}_t|P^{(i)}_{t-1} = \text{Perceived} &\sim \text{Categorical}([0.2, 0.8]) \\		
		F^{(i)}_t|P^{(i)}_t = x &\sim \text{Binomial}(I_\text{Perceived}(x)) \\
		M^{(i)}_t|P^{(i)}_{t-1},O^{(i)},T &\sim \text{Binomial}(\bm{\theta})\\
		\theta_j &\overset{iid}{\sim} \text{Beta}(1,1), \text{ where $j \in [1, 8]$} \\		 
\end{split}
\label{eq:distributions}
\end{align}

The categorical distributions of, $P^{(i)}_t|P^{(i)}_{t-1}$, specify a transition matrix over the perception state. The chosen parameter values for this matrix result in a perception duration governed by a geometric distribution with parameter $p=0.2$, based on the assumption that the player's focus of attention on the victims is about 5 seconds. We found that the model is not very sensitive to this parameter in the range of 3 to 10 seconds, so 5 seconds is a reasonable choice.

Also, to constrain the solution space to valid assignments only, we compute the
conditional distribution of $A$ as:

\begin{equation}
\begin{split}
		A|&O^{(1)},O^{(2)},O^{(3)} = \\ 
		&\begin{cases}
 			\text{P1 got B}, \ \text{ if } O^{(1)} = B, O^{(2)} = O^{(3)} = A \\
 			\text{P2 got B}, \ \text{ if } O^{(2)} = B, O^{(1)} = O^{(3)} = A \\
 			\text{P3 got B}, \ \text{ if } O^{(3)} = B, O^{(1)} = O^{(2)} = A \\
 			\sim \text{Categorical}\left(\frac{1}{3},\frac{1}{3},\frac{1}{3}\right), \ \text{   otherwise}     
    	\end{cases}	
\end{split}
\label{eq:legend_assignment}
\end{equation}

Finally, ToMCAT's belief about the original legend assignment is determined by the posterior $p(A | F^{(i)}_{t\dots \tau}, M^{(i)}_{t\dots \tau})$ for all players $i \in \{1,2,3\}$ and time steps $t \in [0,\tau]$. 

\section{Inference}

We perform approximate inference in the model using a particle filter
algorithm \cite{gordon:1993}. Given the tractability of the substructures in
the model, we apply the Rao-Blackwellised version of this sequential inference
method \cite{murphy:2001, doucet:2003, schon:2005}. Formally, let
$X^{(1:3)}_{0:t}$ indicate all occurrences of player 1, 2 and 3 instances of a
random variable $X$ from time step 0 until time step $t$. The posterior can be
decomposed in the following terms:

\begin{equation}
\begin{split}
	p&\left(P^{(1:3)}_{0:t},O^{(1:3)},T,A|M^{(1:3)}_{0:t},F^{(1:3)}_{0:t}\right) = \\ 
	&p\left(P^{(1:3)}_{0:t}|M^{(1:3)}_{0:t},F^{(1:3)}_{0:t}\right)\overbrace{p\left(O^{(1:3)}|M^{(1:3)}_{0:t},P^{(1:3)}_{0:t}\right)}^{(a)}\\
	&\underbrace{p\left(T|M^{(1:3)}_{0:t},P^{(1:3)}_{0:t},O^{(1:3)}\right)}_{(b)}p\left(A|O^{(1:3)}\right)
\end{split}	
\label{eq:inference}
\end{equation}

We sample each $O^{(i)}$ and $T$ from the analytical form of the posteriors in
$(a)$ and $(b)$ in Eq.~\ref{eq:inference}, and we use particles to approximate the posteriors of the remaining terms. 

\section{Evaluation}

\subsection{Setup}

In the original experiment, each team plays two different missions with
different conditions and configurations of rubble blockages and victims
locations. In this evaluation, we only use the first trial of each team to
avoid introducing bias due to the knowledge the players will develop about the
mission design and also because teams are more likely to bring the collective
marker meaning they adopted in the first trial to the second, which would make
the inference task extremely hard or even impossible. 

We use 52 mission trials to train and evaluate our model using
cross-validation. Independent human observers annotated 12 extra trials at three different
points in time (180, 480 and 780 seconds after the beginning of the mission). We use this trial set to compare our agent's performance with
that of a human observer. 

We use the last predictions as the final answer from
the human observers, which should be their most confident output. Similarly, we use our agent's predictions at time step 780. This guarantees that both ToMCAT and human agents have been exposed to the same evidence up to the decision point. Further, the human observers also know that only one player received a different legend, which is a constraint embedded in our model formulation.

We trained $\bm{\theta}$ in Equation~\ref{eq:distributions} using Gibbs sampling for 600 iterations, from which we discard the first 100 as part of the burn-in phase. Additionally, we used 5000 particles for inference. 

\subsection{Results}

ToMCAT is a probabilistic agent, which means that the confidence of its beliefs
will depend on how likely the given evidence explains specific hidden states.
We transform the predictions from different human observers into probabilities
to compare them against our agent's predictions at different confidence levels.
We define the prediction confidence of the human agent such that if for a given
trial, all observers agreed on a given legend, the human agent is 100\%
confident about that legend. On the other hand, if only two humans agreed on
that legend, the human agent is 2/3 confident about it and 1/3 confident about
the legend chosen by the third human observer. 

\begin{figure}[!thb]
  \centering
  \includegraphics[width=\linewidth]{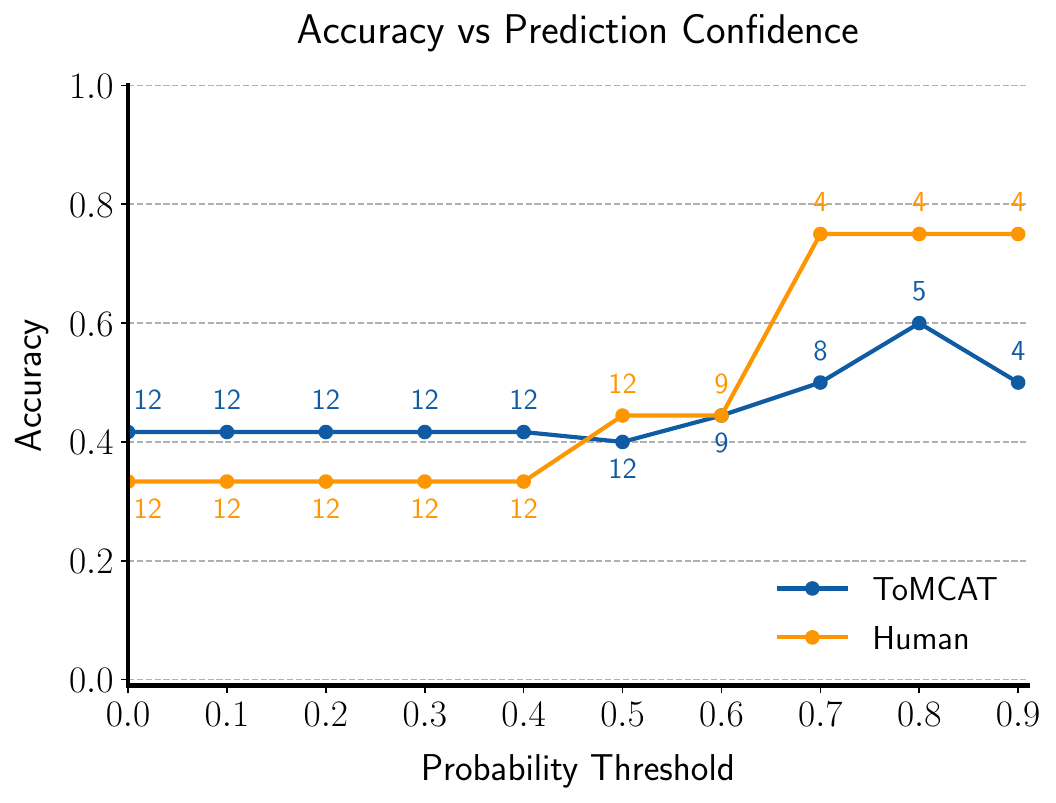}
  \caption{%
    Accuracy of the human and ToMCAT agent at different confidence levels. The
    probability threshold defines the minimum probability that the agent had to
    achieve for the predicted class to count towards the accuracy computation.
    The numbers above and below the lines represent the number of trials for
    which the agent's prediction probabilities were above the defined threshold
    level. The higher the threshold, the smaller the coverage since only the
    most confident predictions are taken into account.
  }
  \label{fig:confidence_levels}
\end{figure}

We compute the agents' performance such that whenever they are not able
to identify a winning legend (uniform probability across legends), we count
that prediction as wrong. We then compare the accuracy of our agent against the
probabilistic human agent at several confidence levels. \figurename \
\ref{fig:confidence_levels} shows that ToMCAT outperforms the human estimations
in the full test set. As we increase the confidence level of the predictions,
the human agent surpasses ToMCAT, but it reduces the coverage of the predicted
trials. Conversely, except for the case in which we only count highly confident
predictions (confidence above 0.9), ToMCAT achieves a reasonable accuracy and
has higher coverage than the human agent.

To ensure that ToMCAT's performance is consistent across different sets, we
performed 5-fold cross-validation on the training set. The results reported in
Table \ref{tab:results} show that ToMCAT can infer false belief significantly
better than chance across different randomly split folds. 

\begin{table}[t]
    \begin{tabular}{lccc}
        \toprule
        & ToMCAT & Chance & Human \\\midrule
        5-Fold CV & $\bm{0.65 \pm 0.17}$ & $0.33$ & Not available\\
        Annotated Set & $\bm{0.41}$ & $0.33$ & $0.33$\\
        \bottomrule
    \end{tabular}
    \caption{%
        Accuracies of ToMCAT, random and human agents at inferring false belief. In
        the first row, the figure for ToMCAT represents the accuracy obtained by
        doing 5-fold cross-validation on the 52 mission trials from the training
        set. This set does not contain human predictions. The second row shows the
        accuracy of the agents in the 12 mission trials from the human-annotated
        test set. Human observers performed no better than random chance on this set.
    }
    \label{tab:results}
\end{table}

Additionally, we are interested in evaluating what premises ToMCAT uses to
update its belief about human teammates. Our experiments show that ToMCAT
developed a consistent theory of the players' minds. ToMCAT correctly adjusts
its belief about the marker meaning adopted by each player as they place
markers in the game given their perception about a nearby victim.
Interestingly, the belief about the legend adopted by the team was also
rationally updated when the agent's uncertainty about which player holds a
false belief changed. These results show that ToMCAT is correctly developing
beliefs about individual and shared mental states.

\begin{figure}[!thb]
  \centering
  \includegraphics[width=\linewidth]{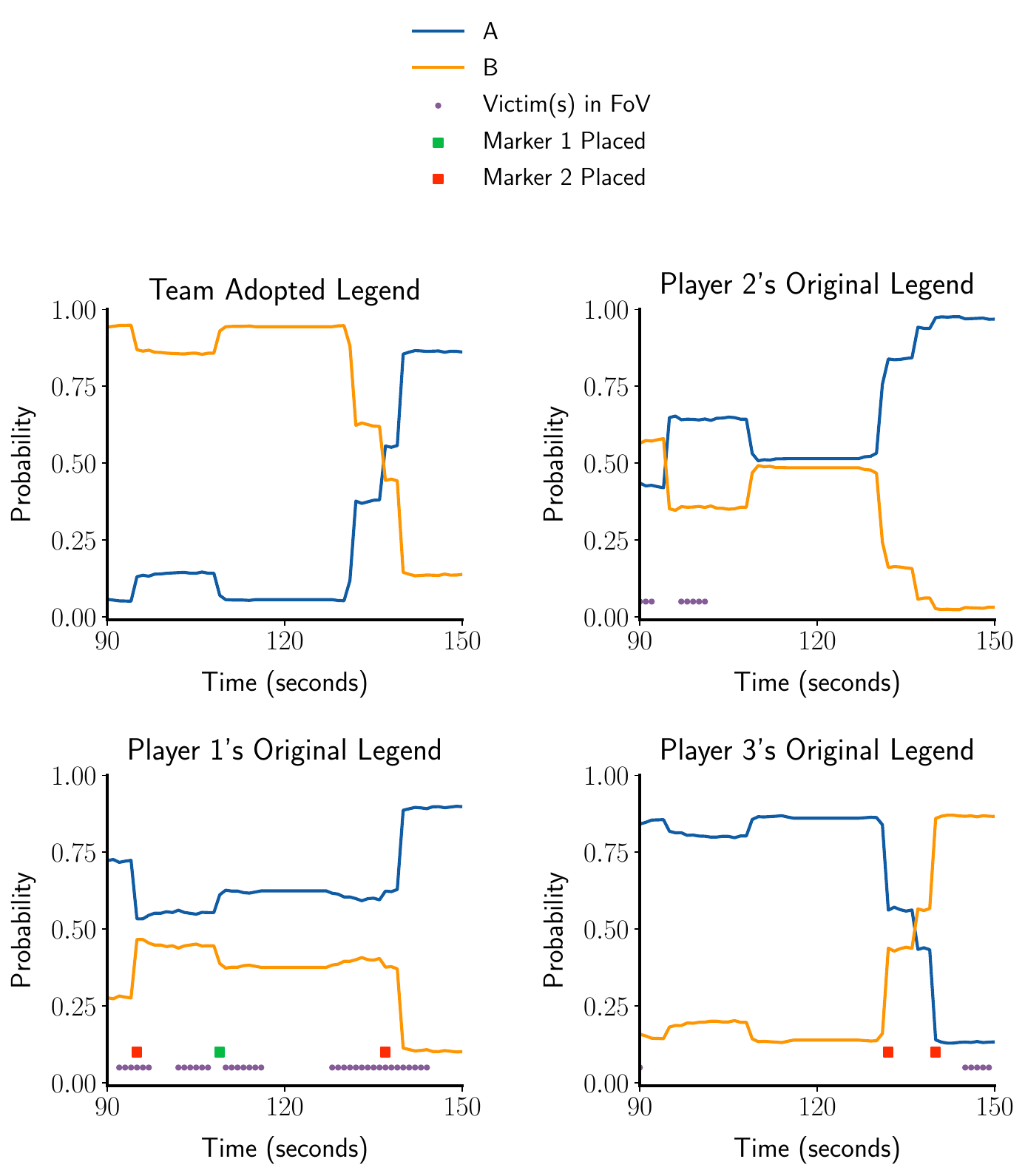}
  \caption{%
    Agent's beliefs about humans mental states being updated over time on a
    sample trial. As players perceive victims and place markers in the building,
    the agent updates its belief about which meaning the players individually and
    collectively are adopting for the markers.
    }
  \label{fig:belief_update}
\end{figure}

\figurename \ \ref{fig:belief_update} shows an example of the agent's beliefs
changing over time for a given trial. Shortly after the first 3 minutes from
the mission start, player 1 places a marker 2 to signal that there are victims
near where the marker is placed, a behavior consistent with legend A. At this
point, the agent does not have much information about this player's actions and
used this evidence to adjust its belief about the team adopted legend at the
expense of incorrectly updating its belief about player 1's legend. However, as
player 1 keeps placing markers according to legend A, the agent can quickly
adjust its belief about this player's original legend.

In the meantime, the agent also observes that player 3 is placing markers
consistent with the meaning provided by legend B. It rapidly infers that the
player holds the false belief and adjusts its belief about the player's and
team's legends accordingly. 
    
\begin{figure}[!thb]
	\centering
	\begin{subfigure}{.5\textwidth}
	    \centering
		\includegraphics[width=0.9\linewidth]{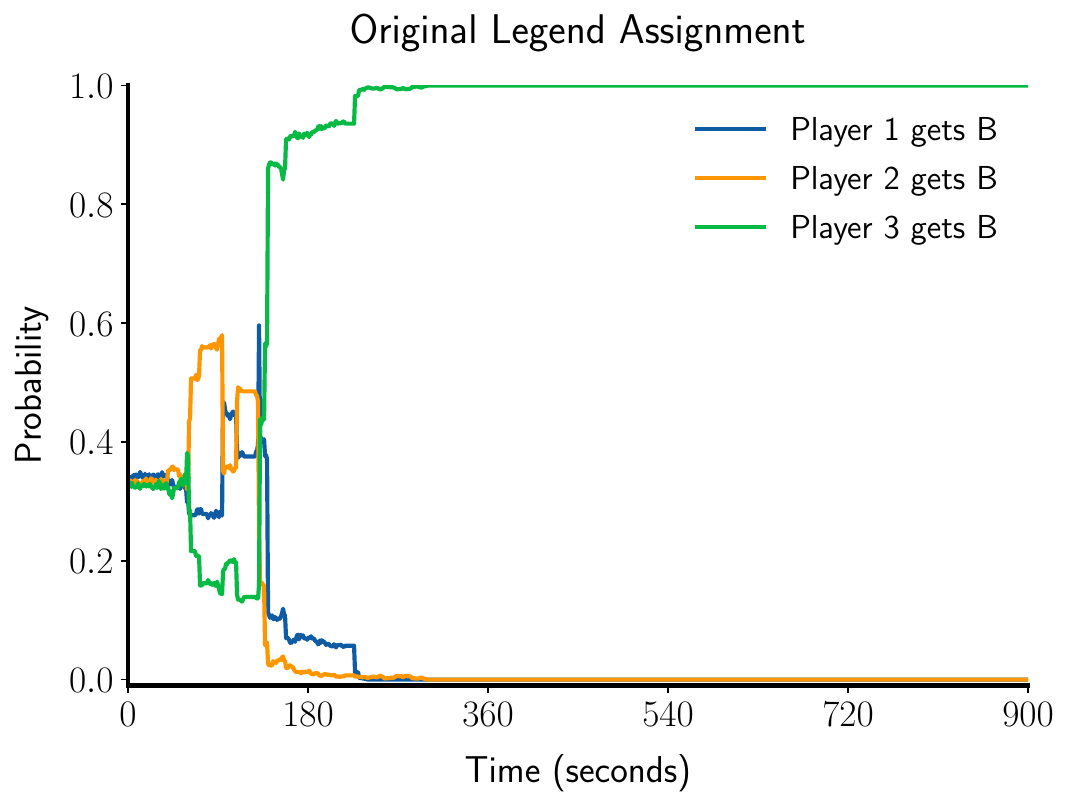}
		\caption{Confident belief}
		\label{fig:confident_belief}		
	\end{subfigure}
	\begin{subfigure}{.5\textwidth}
	    \vspace{1em}
		\centering
		\includegraphics[width=0.9\linewidth]{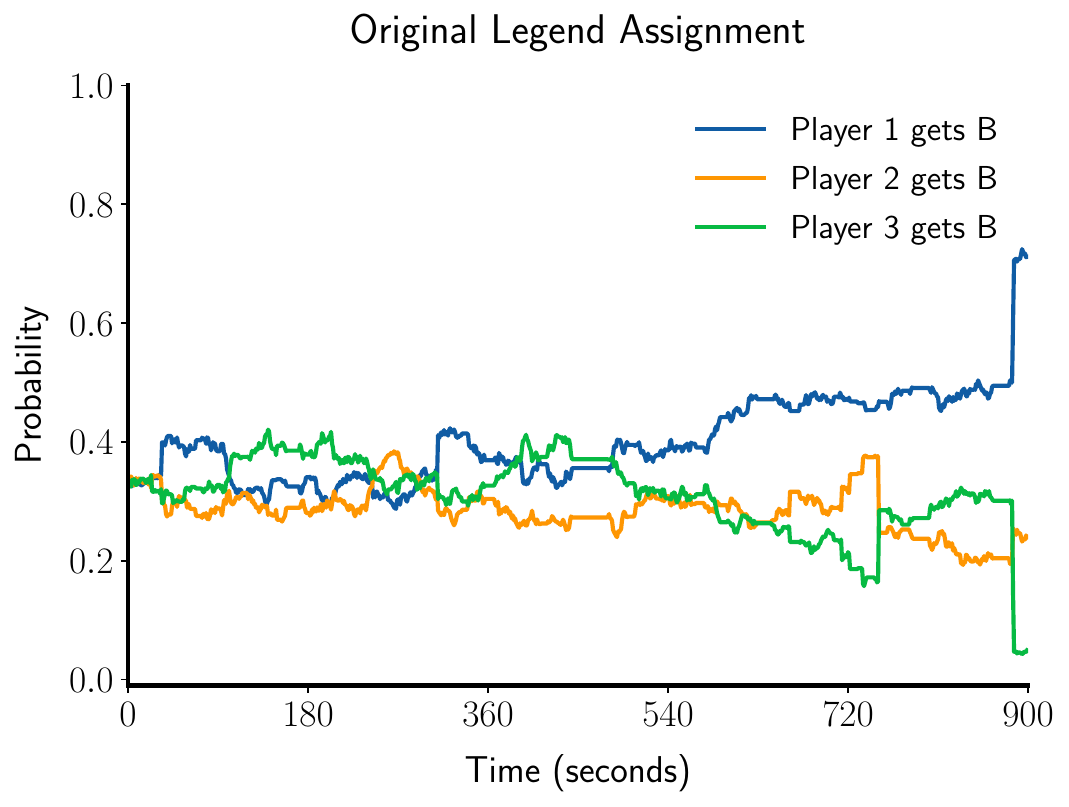}
		\caption{Uncertain belief}
		\label{fig:uncertain_belief}
	\end{subfigure}
	\caption{%
        (a) An example of a trial in which players are active at placing markers.
        ToMCAT's beliefs quickly adapts and it becomes very certain about the correct
        original assignment. (b) An example of a trial where players do not use the
        markers very often. ToMCAT is not very certain until the end of the mission
        when player 3 places a marker consistent with legend A, which increases ToMCAT
        current believe that player 1 is the one that holds the false belief.
    }
	\label{fig:complete_belief_update}
\end{figure}

\figurename \ \ref{fig:complete_belief_update} illustrates how the agent changes
its belief about original assignments across a whole trial. The plots
illustrated that the agent is not always confident about its belief, but it can
quickly adapt whenever it perceives a particular behavior.  

We identified that the scenarios in which the model developed a wrong belief
about the legend assignments were mostly because players barely or never used
markers or because players created a different use for them. In one instance, a
player used marker 2 to keep track of rooms with rubble that needed to be
cleared. The agent will incorrectly think players are using legend B as there
are no victims around, just rubble.  Another scenario in which the model
develops a wrong belief about the assignments is when players mistakenly
place wrong markers and immediately afterward replace them with correct ones. A
final explanation would be the strong assumption we made that players perceive
all the victims in their fields of view. The majority of these issues are
tractable with simple improvements in the model, and we will address them in
future work.

\section{Conclusion}

Our experiments showed that ToMCAT was able to develop a rational theory of
human minds to accurately identify false beliefs in the context of a search and
rescue mission. Our agent explained players' behaviors not only by their states
but also by a collective, shared state. Importantly, it is able to infer false
beliefs for some trials with high levels of confidence (e.g., the one in
Figure \ref{fig:complete_belief_update}), thus making it a viable proposition to
intervene at these points (e.g., approximately 200 seconds into the mission
shown in Figure \ref{fig:confident_belief}.) to correct false beliefs and
misaligned mental models.

ToMCAT performed significantly better than chance and achieved results comparable with human predictions. For future works, we will increase the number of human-annotated trials and explore different information asymmetry scenarios. ToMCAT knows that only one of the players received a different legend. However, there are cases in real life in which the form of information asymmetry cannot be determined beforehand, and the agent will need to infer that.

Finally, players are encouraged to communicate in the mission, and their speech
is run through our automatic speech recognition (ASR) and rule-based information extraction pipelines. But the
current agent does not ground its belief in this kind of information. We can
likely improve ToMCAT's performance by using the outputs of our natural language
processing systems as evidence to infer human mental states.

\section{Acknowledgments}

This research was conducted as part of DARPA's Artificial Social Intelligence
for Successful Teams (ASIST) program, and was sponsored by the Army Research
Office and was accomplished under Grant Number W911NF-20-1-0002. The views and
conclusions contained in this document are those of the authors and should not
be interpreted as representing the official policies, either expressed or
implied, of the Army Research Office or the U.S. Government. The U.S.
Government is authorized to reproduce and distribute reprints for Government
purposes notwithstanding any copyright notation herein.

\bibliography{bibliography}

\end{document}